\documentclass{article}
\usepackage[preprint]{neurips_2019}
\PassOptionsToPackage{numbers, compress}{natbib}
\usepackage[utf8]{inputenc} 
\usepackage[T1]{fontenc}    
\usepackage{hyperref}       
\usepackage{url}            
\usepackage{subcaption}
\usepackage{booktabs}       
\usepackage{amsfonts}       
\usepackage{nicefrac}       
\usepackage{microtype}      
\usepackage{graphicx}
\usepackage{amsmath}
\usepackage{bm}

\usepackage{amssymb}
\usepackage{xcolor}
\usepackage{multirow}
\usepackage{nicefrac}

\usepackage[compress]{natbib}
 
\usepackage[ruled,vlined]{algorithm2e}

\title{Patient-Specific Effects of Medication \\Using Latent Force Models with Gaussian Processes}

\author{
   Li-Fang Cheng \\
   Princeton University\thanks{Now at Verily Life Sciences, LLC. All work was done at Princeton University.}\\
   \texttt{lifangc.cheng@gmail.com} \\
   \And
   Bianca Dumitrascu \\
   Princeton University\\
   \texttt{bidumit@alum.mit.edu} \\
   \And
   Michael Zhang \\
   Princeton University\\
   \texttt{mz8@cs.princeton.edu} \\
   \And
   Corey Chivers \\
   University of Pennsylvania Health System \\
   \texttt{corey.chivers@uphs.upenn.edu} \\
   \And
   Michael Draugelis \\
   University of Pennsylvania Health System \\
   \texttt{michael.draugelis@uphs.upenn.edu} \\
   \And
   Kai Li \\
   Princeton University\\
   \texttt{li@cs.princeton.edu} \\
   \And
   Barbara E. Engelhardt \\
   Princeton University\\
   \texttt{bee@princeton.edu} \\
}

\begin{document}
\maketitle

\begin{abstract}
Multi-output Gaussian processes (GPs) are a flexible Bayesian nonparametric framework that has proven useful in jointly modeling the physiological states of patients in medical time series data. However, capturing the short-term effects of drugs and therapeutic interventions on patient physiological state remains challenging. We propose a novel approach that models the effect of interventions as a hybrid Gaussian process composed of a GP capturing patient physiology convolved with a latent force model capturing effects of treatments on specific physiological features. This convolution of a multi-output GP with a GP including a causal time-marked kernel leads to a well-characterized model of the patients' physiological state responding to interventions. We show that our model leads to analytically tractable cross-covariance functions, allowing scalable inference. Our hierarchical model includes estimates of patient-specific effects but allows sharing of support across patients. Our approach achieves competitive predictive performance on challenging hospital data, where we recover patient-specific response to the administration of three common drugs: one antihypertensive drug and two anticoagulants.
\end{abstract}

\section{Introduction}

In the era of digital medicine, modern medical devices enable clinicians to accurately and frequently measure the physiological state of their patients. Heart rate, blood pressure, arterial oxygen saturation, and white blood cell counts are just a few of the vital signs used to monitor patient state and to predict outcomes such as hospital discharge rate, patient survival, or readmission~\citep{rajkomar2018scalable,ranganath2016deep}. 
To model the complex relationships governing patient state dynamics, a large body of work takes advantage of the flexible properties of Gaussian processes (GPs). Both single-output \citep{stegle2008gaussian,lasko2013computational} and multi-output methods \citep{discover-cardio-dynamics-2012,ghassemi2015multivariate,cheng2017arXiv,futoma2017improved} use carefully designed kernels to improve prediction accuracy of patient states across sparse, noisy, and irregularly sampled time series horizons. These methods predict correlations between patient vital signs that give insight into imputing hard-to-sample state variables, detecting patient trends, and making decisions under uncertainty. 
Modeling patient state dynamics is made difficult by the administration of drugs and other therapeutic interventions, which directly impact state features for a limited time window and often return to baseline.  Accurate prediction of patient state following an intervention allows a clinician to consider multiple options for intervention and make informed treatment decisions.

Recent papers have attempted to model the complex effects of interventions on patient state \citep{Xu2016-bayesian,schulam2017counterfactual}. These models do not incorporate a control systems understanding of the complex mechanisms governing the tightly controlled responses of physiological traits to interventions. This leads to inaccurate and non-generalizeable predictions when insufficient training data is available, which is often the case when predictions include a patient-specific component.

One motivation for modeling intervention effects on patient state with dynamical systems is to learn optimal treatment policies. For instance, prior work adapted reinforcement learning (RL) to derive a closed-loop anesthesia controller to regulate mean arterial pressure based on a dynamical patient model \citep{padmanabhan2015closed}. Other applications, such as finding multi-drug therapies for human immunodeficiency virus (HIV), have used dynamical systems models \citep{adams2004dynamic}.

In this paper, we introduce a Bayesian nonparametric framework for estimating the dynamics of clinical traits in response to drug interventions through electronic health records (EHRs) from hospital patients. Specifically, we develop an approach to learn the patient-specific response of clinical traits to treatments from medical time series data. Our approach consists of a multi-output Gaussian process (GP) derived from latent force models (LFMs) \citep{alvarez2009latent}, which we model using GPs with kernels derived from differential equations representing dynamical systems. 
To model the effect of multiple treatments on patient-specific clinical traits, we use latent force functions sampled from Gaussian processes with \emph{causal kernels}. 
By estimating patient-specific parameters to capture the effects of treatments on specific clinical traits, our model offers novel mechanistic insights and achieves competitive predictive accuracy of physiological response to interventions when compared with state-of-the-art methods \citep{soleimani2017treatment}.

The contributions of this work are two-fold. First, we compose causal kernels in time-marked medical data to capture the latent dynamics of interventions effects using a GP. Second, we incorporate these treatment effects in a model of patient state by convolving this causal kernel latent force model with a multi-output GP that captures patient state absent interventions. Our approach is a necessary step towards achieving optimal control of patient health through targeted, personalized treatments \citep{sarkka2017gaussian}. We show the predictive value of our approach on a large hospital data set.

\section{Background}

\subsection{Gaussian Processes for Time Series}
In the medical time series setting, data are collected from $n$ patients, indexed by $i$, across $T_i$ time points indexed by $t$. Single output data typically correspond to time-varying covariates encoding physiological states such as blood pressure or heart rate. The measured pairs $\mathcal{D} = \{x_{i}, y_{i}\}_{i=1}^n$, where $x_{i}$ corresponds to the time at which the covariate $y_{i} \in \mathbb{R}$ was recorded, are then modeled through an underlying latent function $f(\cdot)$ such that $y_i = f(x_i) + \epsilon_i$, where $\epsilon_i \sim \mathcal{N}(0, \sigma_i^2)$ is independent Gaussian noise. The goal is to estimate and to evaluate $f(\cdot)$ at future time points to enable prediction and early detection of physiologically abnormal states. 

Gaussian processes (GPs) represent a
versatile generative framework for modeling the distribution on a real-valued function $f(\cdot)$. GPs are nonparametric stochastic processes specified by mean and covariance functions:
\begin{equation}
    f(\mathbf{x}) \sim \mathcal{GP}(\mu(\mathbf{x}), k(\mathbf{x}, \mathbf{x}')),
\end{equation}
where $\mu(\mathbf{x})$ is the \emph{mean function} 
$\mu(\mathbf{x}) = \mathbb{E}[f(\mathbf{x})]$ 
and $k(\mathbf{x}, \mathbf{x}')$ is the \emph{covariance function} or \emph{kernel}:
$k(\mathbf{x}, \mathbf{x}') = \mathbb{E}[(f(\mathbf{x})-\mu(\mathbf{x}))(f(\mathbf{x}')-\mu(\mathbf{x}'))]$.
The mean function $\mu(\mathbf{x})$ is often assumed to be zero \citep{Rasmussen2006}. An immediate result is the multivariate Gaussian form of the joint $[f(x_1),f(x_2), \ldots f(x_n)] \sim \mathcal{N}(\mathbf{0}, K)$, where $K$ is the $n \times n$ kernel matrix with entries $K_{i,j} = k(x_i, x_j)$. 

Properties of the function $f(\mathbf{x})$ such as smoothness or periodicity are determined by the kernel function $k(\mathbf{x}, \mathbf{x}')$. One commonly used kernel is the squared exponential (SE) kernel
\begin{equation}
k(\mathbf{x}, \mathbf{x}') = \sigma^{2}\exp{\left( -\frac{||\mathbf{x}-\mathbf{x}'||^{2}}{2\ell^{2}} \right)},
\end{equation}
which is parameterized by a length scale $\ell$ and a scale factor $\sigma$. The functions generated by a GP with an SE kernel are smooth because the kernel function is infinitely differentiable. SE kernels capture \emph{stationary} processes, as the covariance between two vectors depends on the difference in time (or other covariate) $||\mathbf{x}-\mathbf{x}'||$ but not on absolute time.

GPs have been studied extensively in the context of time series \citep{Roberts20110550}, and are especially useful when the data are sparsely or irregularly sampled. In medical time series, the clinical measurements are recorded sporadically and sometimes sparsely across time. To better model multiple correlated measurements, multi-output GPs (MOGPs), which capture the covariance structure between multiple measurements, have been adapted for use on medical data \citep{ghassemi2015multivariate,futoma2017improved,cheng2017arXiv}. However, kernels used to capture temporal dependencies between sequential clinical measurements are mostly stationary. This is a limitation since, during a patient's stay in the hospital, clinical events and interventions occur, and physiological dynamics often vary across time.

\subsection{Latent Force Models for Patient Data}
Physiological dynamics have been long studied by physiologists using systems of differential equations. In the case of heart rate and blood pressure, for example, the cardiac conduction system is assumed to be a network of self-excitatory pacemakers leading to systems of nonlinear oscillators \citep{glass2001synchronization}. Medical time series-based prediction of such covariates has relied on regression models, 
linear and nonlinear, including but not limited to GPs; yet an explicit connection with control theory has been lacking. Recently, several methods were proposed to bridge the gap between stochastic control methods and nonparametric time series models \citep{gao2008gaussian,alvarez2009latent}. 
In particular, multi-output GPs are one approach to representing latent force models, where the covariance functions (kernels) are derived using ordinary differential equations (ODEs). In the LFM setup, we would like to model the system dynamics between a set of observed processes, $\lbrace g_q(t) \rbrace_{q=1}^{Q}$, and a set of unobserved \emph{latent forces}, $f_m(t)$, assuming that they interact according to differential equations capturing those dynamics. For example, in a first-order LFM, the following equation holds:
\begin{equation}
    \frac{\text{d} g_q(t)}{\text{d}t} + D_q g_q(t)= B_q + \sum_{m=1}^{M}S_{qm} f_m(t),
\end{equation}
with $B_q$ and $D_q$ representing the base level value and underlying decay parameter of each output $q$, and $S_{qm}$ the influence constants from each latent force $f_m$ to each output $g_q(t)$. Through this formulation, one can recover the latent force functions $f_m(t)$ and the output functions $g_q(t)$ from discrete observations. Solving the ODE yields
\begin{equation}
    g_q(t) = \frac{B_q}{D_q} + \sum_{m=1}^{M}S_{qm} \exp(-D_q t) \int_0^t f_m(\tau) \exp(D_q \tau) \text{d}\tau.
    \label{eq:lfm_integral}
\end{equation}
In LFMs, the latent forces $f_m$ are modeled independently as samples from their respective GPs. For certain classes of covariance functions, such as SE kernels, one can show that the outputs $g_q$ are also GPs with analytically closed-form covariance functions, as well as cross-covariances between latent forces and the outputs \citep{alvarez2009latent}.

While sampling the latent forces $f_m$ from independent GPs is computationally convenient, important information can be lost. For example, when provided with historical patient data, we might want to know how latent physiological dynamics are shared across patients from related covariate groups (i.e., same age, sex, BMI) or across patients receiving similar treatments. Limited numbers of observations also motivates a hierarchical approach to this problem, in order to share strength across patients.

\subsection{Treatment Effect Estimation}
In addition to modeling physiological outputs, accurate modeling of medical data sets requires incorporating various treatment effects, with the purpose of controlling or stabilizing the physiological states of patients. Treatments are often drug interventions that can be characterized by drug name, administration type (e.g., oral, intravenous), and dosage. Estimating the effect of a treatment on a patient's physiological state is paramount to improving prediction and evaluating treatments.

While GPs are commonly used to model medical time series data, several extensions have been proposed to estimate the effects of medical treatments. Counterfactual Gaussian processes (CGPs) \citep{schulam2017counterfactual} use marked point processes (MPP) as an event model to account for dependencies between actions and observed physiological trajectories. In \cite{Xu2016-bayesian}, a class of parametric functions is introduced to model the effects of dialysis for patients with acute kidney injury. The functions were designed to explicitly model different types of effects, including delay and decay. To explain heterogeneity across patients, a Dirichlet process was used for clustering patients. In \cite{futoma2017improved}, the treatment effects are modeled in the prior mean function of multi-output Gaussian processes, formulated as the sum of multiple exponential decay functions. These approaches require the response dynamics to conform to a specific functional form such as exponential decay, where in practice they are much more heterogeneous. 

More recently, \cite{soleimani2017treatment} introduces treatment effects as the output of a linear time-invariant (LTI) system. The inputs are the observed administration of medications (e.g., type and dosage), and the effects are estimated based on a chosen form of a second-order filter. Patient-specific filter parameters were estimated and regularized using a global prior across patients. 

To allow the treatment response to take on arbitrary functional forms with scalable effects sizes and directions, we use a hierarchical GP to model the latent forces. In particular, we replace the independent GPs with a causal-kernel GP whose hyperparameters are shared across patients to allow arbitrary functional response and to share strength by representing correlations across patient groups.

\section{Causal Convolutional Gaussian Processes}

Here, we propose a flexible framework to model medical time series data. Our method brings together ideas from GP latent force models \citep{alvarez2009latent,sarkka2017gaussian} and causal GPs \citep{pmlr-v22-cunningham12} to address a  challenge in modeling medical time series data---\textit{the systematic inclusion of multi-treatment effects on the dynamics of multiple physiological covariates}. We first introduce the notation in the context of medical time series data, and then introduce the model. We also discuss the details of implementation and inference methods.

\subsection{Medical Time Series with Treatments}
We denote the observed medical time series data $y_{i,j}(t)$ from $i=\{1,2,\ldots, n\}$ patients characterized by $j=\{1,2,\ldots, J\}$ physiological dynamic covariates across irregularly sampled time points $t=\{1,2,\ldots, T_{i,j}\}$, as noisy samples from a Gaussian process with a patient- and covariate-specific mean function $\mu_{i,j}(t)$ and kernel $k^b_{i,j} (t,t')$:
\begin{align}
    y_{i,j}(t) &= f_{i,j}(t) + \epsilon_j, \mbox{    }\epsilon_j \sim \mathcal{N}(0, \sigma_{i,j}^2) \nonumber\\
    f_{i,j}(t) &\sim \mathcal{GP} \left(\mu_{i,j}(t), k^b_{i,j} (t,t') \right).
\end{align}
Here, the kernel $k^b_{i,j}$ accounts for stationary temporal fluctuations of physiological signals, such as circadian rhythms. We choose this kernel as a sum of one SE kernel and one periodic kernel:
\begin{align}
     k^b_{i,j} = k_{\textsc{Se}} + k_{\textsc{Per}} = \sigma_{1,i,j}^2 \exp \left[ -\frac{(t-t')^2} {2\ell^2_{1,i,j}} \right] + \sigma_{2,i,j}^2 \exp{ 
              \left[
              -
              \frac{
              \sin^{2}\left(\nicefrac{\pi||t-t'||}{p_{i,j}}
              \right)
              }{2\ell_{2,i,j}^{2}}  
              \right] }.
\label{eq:baseline_kernel}
\end{align}
For each patient $i$, we index treatments given as $m=1,2, \ldots, M_i$, and the time of treatments is denoted as $t^i_1, t^i_2,\ldots t^i_{M_i}$. For each treatment, we use function $\tau_i : {1,2, \ldots,M_i} \to \mathcal{T}$ to map the index of the treatment to a treatment set $\mathcal{T}$ representing the type of treatment. Since the dosage-response curve of the same drug usually has a nonlinear curve that varies across dosages \citep{myers1980metoprolol,ghassemi2014data}, and the characteristics of absorption vary across different routes, we treat the same drug with different dosages or taken via different routes (e.g., oral or injection) as different treatments. Whenever clear from the context, we drop the patient index from these variables. 

We assume there are different latent forces induced by each type of treatment. For a treatment $m$ given at time $t_{m}$, we model the latent force as a function of time $f_{m}(t; t_{m})$ drawn from a Gaussian process. We also assume that each patient has a patient-specific latent force, and use a hierarchical model to share support for latent force models across patients.

\subsection{Causal Treatment Dynamics}
The dynamic behavior of a treatment's response to the clinical covariates are represented in our setup as a latent force model. In particular, we model the mean function $\mu_{i,j}(t)$ of the clinical traits through the first-order dynamical system LFM:
\begin{align}
\frac{\text{d} \mu_{i,j}(t)}{\text{d}t} + D_{i,j}\mu_{i,j}(t) = B_{i,j} + \sum_{m=1}^{M_{i}} S_{i,j,m} f_{m}(t;t_m),
\end{align}
where the decay $D_{i,j}$, the baseline covariate output $B_{i,j}$, and treatment effect size $S_{i,j,m}$ are patient-specific parameters that control the dynamics of the treatment response. We assume these patient-specific parameters come from a population-wise empirical prior based on demographic data, such as age and weight.
We assume the latent force function $f_m(t;t_m)$ of the same treatment is shared across patients, and is sampled from a Gaussian process:
\begin{align}
    f_m(t;t_m) \sim \mathcal{GP} (0, k_{f,f'}(t,t'; t_m)).
\end{align}

We introduce causality on the latent force functions associated with each treatment, which means that causal effects must only act forward in time. To do this, we designed the kernel $k_{f,f'}(t,t'; t_m)$ as a causal kernel~\citep{pmlr-v22-cunningham12}. That is, 
\begin{align}
    k_{f,f'}(t,t'; t_m) = \exp \left\{ - \frac{[h(t-t_m) - h(t'-t_m)]^2}{\ell_m^2} \right\},
\end{align}
where $h(t) =  t \mathcal{I}(t>0)$ is the clipping function warping the input space and enforcing causality, while preserving the GP structure \citep{pmlr-v22-cunningham12}. Through $h(t)$, the function $f_m(t;t_m)$ is constant before the current time $t_{m}$. In our model of treatment effects, we introduce an additional condition $f_m(t;t_m) = 0$ for $t < t_m$.

\subsection{Kernel Convolution}
The structure of the latent force model leads to a natural composition with the causal Gaussian process prior, leading to an analytic computation of output covariances and cross-covariances. This fact allows for simple gradient descent-based inference. Closed-form kernels were derived following the integral in Eqn. \ref{eq:lfm_integral}. For instance, the cross-covariance between $\mu_{i,j}(t)$ and one latent force $f_{m}(t)$ when $t, t' > t_{m}$ is computed as
\begin{equation}
\begin{array}{rll}
&k_{\mu_{i,j},f_{m}}(t,t';t_{m})= \displaystyle S_{i,j,m}\exp{(-D_{i,j}t)}\exp{\left[
    -\left(
    \frac{
    t'-t_{m}
    }{\ell_{m}}
    \right)^{2}
    \right]}
 \displaystyle \times 
    \frac{1}{D_{i,j}}\left[
    \exp{(D_{i,j}t_{m})}-1
    \right]
    &\\\\
&+ \displaystyle  \frac{\sqrt{\pi}\ell}{2}S_{i,j,m} \exp{[-D_{i,j}(t-t')]} \exp{(\nu_{i,j,m}^{2})} \displaystyle \times
\left[ 
\text{erf}\left( 
\frac{t-t'}{\ell_{m}}-\nu_{i,j,m}
\right)
+
\text{erf}\left(
\frac{t'-t_{m}}{\ell_{m}}+\nu_{i,j,m}
\right)
\right],
& \\
\end{array}
\end{equation}
where $\nu_{i,j,m}=\frac{\ell_{m}D_{i,j}}{2}$. Details of the computation of the closed-form kernels corresponding to the cases 
$t> t_{m}>t'$, $t_{m}>t, t'$, and $t'> t_{m}>t$ are in the Supplementary Material.

\subsection{Hyperparameter Learning for Medication Effects Model}

To learn the hyperparameters for each patient, we optimize the marginal likelihood of the GP model with respect to the vector of observations across covariates $\mathbf{x}_{i}$ and $\mathbf{y}_{i}$.
\begin{equation}
\label{equation:GP_marginal_likelihood}
\begin{array}{ll}
\log{p(\mathbf{y}_{i}|\mathbf{x}_{i}, \bm{\theta})}= -\frac{1}{2}(\mathbf{y}_{i}-\bm{\mu}_{i})^{\top}(K_{i|\bm{\theta}}+\bm{\epsilon}I)^{-1}(\mathbf{y}_{i}-\bm{\mu}_{i}) \\
\quad\quad\quad\quad - \frac{1}{2}\log{|K_{i|\bm{\theta}}+\mathbf{\epsilon}I|} - \left(\frac{\sum_{j=1}^{J}T_{i,j}}{2}\right)\log{(2\pi)},\\
\end{array}
\end{equation}
where $\bm{\mu}_{i} = \mu_{i,j}(\mathbf{x}_{i})$. For each patient, we learn a set of hyperparameters $\bm{\theta}$ for the baseline kernel for each covariate $\lbrace \sigma_{1,i,j},\ell_{1,i,j},\sigma_{2,i,j},\ell_{2,i,j},p_{i,j}\rbrace$, and the hyperparameters of the derived causal LFM kernel, with the mean function sampled from $\lbrace B_{i,j},D_{i,j},S_{i,j,m},\ell_{i,j,m}\rbrace$. 
We assume that the patient- and treatment-specific hyperparameters are shared across treatments of the same type, while the treatment effect size parameter differs for different dosages or modes of administration. Our implementation is based on GPy \citep{gpy2014}, and we optimize the hyperparameters using scale conjugate gradient methods. We derived the gradients using the SymPy package \citep{sympy}. 

\section{Experiments}
In this section, we show the effectiveness of our method by modeling multiple treatment effects on EHR data from the Hospital of University of Pennsylvania (HUP). We briefly describe the data and preprocessing procedures, and then we discuss results from our method fitted to patient subsets motivated by two clinical applications. We show empirical results of our method using the metrics of prediction accuracy. We compare results with baseline methods with GPs using basic kernels and mean functions from related work~\citep{futoma2017improved,soleimani2017treatment}.

\begin{table}[t]
\begin{subtable}{.5\linewidth}
\begin{tabular}{ll}
\toprule
Antihypertensive Agents ($N=181$) & Count \\
\midrule
Heart Rate (HR) & 9,798 \\
Systolic Blood Pressure (SBP) & 7,804 \\
Metoprolol Tartrate (6.25 mg) & 42 \\
Metoprolol Tartrate (12.5 mg) & 92 \\
Metoprolol Tartrate (25 mg) & 93 \\
Metoprolol Tartrate (50 mg) & 40 \\
Metoprolol Succinate ER (25 mg) & 22\\
Metoprolol Injection (5 mg) & 21 \\
\bottomrule
\end{tabular}
\end{subtable}%
\begin{subtable}{.5\linewidth}
\begin{tabular}{ll}
\toprule
Anticoagulants ($N=404$) & Count \\
\midrule
Partial Thromboplastin Time (PTT) & 4,911 \\
International Normalized Ratio (INR) & 4,348\\
Heparin Injection (5000 units) & 246 \\
Heparin Infusion (25000 units) & 319 \\
Warfarin (5mg) & 27 \\
\bottomrule
\end{tabular}
\end{subtable}
\vspace{2mm}
\caption{\textbf{Data statistics of the two drug types used in the experiments.} Total number of observations for the targeted vital signs and lab results, and the count of targeted treatments.}
\label{table:data_statistics}
\end{table}

\subsection{Inpatient Hospital Data}
We evaluated our method using clinical data collected at HUP. The data set consists of 139k patients with access to demographic details (e.g., age, weight, gender), as well as 139 clinical measurements consisting of vital signs and lab tests, and administrated medications during the patients' stay in the hospital. We normalized each clinical trait by subtracting the empirical mean for each patient from each measurement. We tested our method on two challenging applications---modeling the effects of antihypertensive agents and anticoagulants. We chose to focus on the patients with a primary diagnosis of myocardial infarction (MI; i.e., heart attack) in our data set, resulting in total 1,716 adult patients, as they usually received both types of treatments. 

We first modeled the effects of the most frequently administered antihypertensive drug in our data set, metoprolol, on heart rate (HR) and systolic blood pressure (SBP). Metoprolols are beta-blockers that are mainly used to treat high blood pressure or angina due to heart disease. We filtered MI patients to include patients with at least five observations in both heart rate or blood pressure, reducing the number of patients to 594. We removed patients that were administrated metropolol jointly with other antihypertensive agents, resulting in 233 patients. Finally, we included treatments of metoprolol that were administered at least 20 times across all patients, resulting in 181 patients with six type of treatments, including four different dosages of metoprolol tartrate, and one dosage for metoprolol succinate ER and metoprolol injection, and a total of 310 treatment administrations.

Second, we modeled the effect of two different types of anticoagulants: heparin and warfarin. We filtered the MI patients to include those with at least five observations on two lab test results that reflect a patient's ability to form blood clots: partial thromboplastin time (PTT) and international normalized ratio (INR), resulting in 581 patients. Among all treatments, we considered the top three most frequently administered, and we include patients that received at least one of them. The filtered data includes 404 patients with a total of 592 treatment administrations (Table \ref{table:data_statistics}).

\subsection{Evaluation Metrics}
We evaluated our model by comparing predictive performance with three state-of-the-art GP models: (i) univariate GPs with a constant mean function and the baseline (squared exponential and periodic) kernel $k_{i,j}^{b}(t, t')$ (Eqn. \ref{eq:baseline_kernel}; \textsc{Se+Per}), (ii) univariate GPs with an exponential decay mean function and the Ornstein-Uhlenbeck (OU) kernel \citep{futoma2017improved} (\textsc{Ou+Exp}), and (iii) Mat$\acute{e}$rn-$\nicefrac{3}{2}$ kernel with a second-order LTI filter for effect modeling \citep{soleimani2017treatment} (\textsc{Mat32+Lti}). Note that for the comparative method (ii) and (iii), we added a constant component in the mean function in the proposed setup to account for patient-specific baseline values of each covariate. For all methods, we used the first 70\% of observations for each patient for training, and the remaining 30\% for testing. We computed the mean absolute error (MAE) of predictions on test data to evaluate model performance.

\begin{table*}[t!]
\begin{center}
\begin{tabular}{lcccc}
\hline
\multirow{2}{*}{Covariate} & \multicolumn{4}{c}{Antihypertensive Agents}\\
\cline{2-5}
& \textsc{Se+Per} & \textsc{Ou+Exp} & \textsc{Mat32+Lti} & \textbf{Proposed} \\
\hline
SBP & $10.688 \pm 0.242$ & $11.547 \pm 0.292$ & $\mathbf{10.502 \pm 0.240}$ & $10.882 \pm 0.242$\\
HR & $ \mathbf{7.694 \pm 0.173}$ & $7.780 \pm 0.186$ & $7.791 \pm 0.173$ & $7.851 \pm 0.179$\\
\hline
\multirow{2}{*}{Covariate} & \multicolumn{4}{c}{Anticoagulants}\\
\cline{2-5}
& \textsc{Se+Per} & \textsc{Ou+Exp} & \textsc{Mat32+Lti} & \textbf{Proposed} \\
\hline
PTT & $ 12.646 \pm 0.379$ & $12.431 \pm 0.512$ & $12.436 \pm 0.376$ & $\mathbf{12.248 \pm 0.368}$\\
INR & $0.204 \pm 0.009$ & $0.339 \pm 0.013$ & $0.596 \pm 0.011$ & $\mathbf{0.180 \pm 0.009}$\\
\hline\hline
\end{tabular}
\caption{\textbf{Prediction results on test data.} MAE ($\pm$ standard error) were computed for results from our method and three comparison methods using univariate GPs with different mean functions and kernels. Our method performs competitively across covariates under the treatments of antihypertensive agents and better than the comparison methods under the treatments of anticoagulants.}
\label{table:numerical_results}
\end{center}
\end{table*}

\begin{figure*}[t!]
    \centering
    \includegraphics[width=10.5cm]{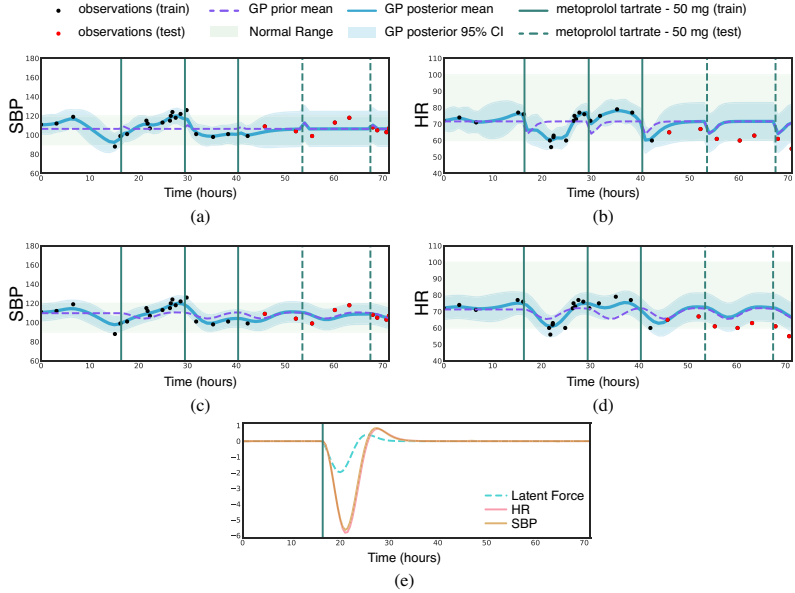}
    \caption{\textbf{Prediction of systolic blood pressure (SBP) and heart rate (HR) for one patient under metroprolol tartrate treatments.} (a) and (b): results from a related method \textsc{Mat32+Lti}. (c) and (d): results from the proposed method. (e) The learned latent force for the first metroprolol tartrate (50 mg) treatment and the effects on SBP and HR; the recovered effects on SBP and HR are in their original units. Our method achieves higher confidence (lower variance) and greater consistency with known clinical effects than the related model. }
    \label{fig:metoprolol_results}
    \vspace{-4mm}
\end{figure*}

\subsection{Clinical Impact}

When comparing the MAE on test data for the two experiments (antihypertensive agents and anticoagulants; Table \ref{table:numerical_results}), our method performed competitively across experiments on the predictive tasks for the covariates responding to antihypertensives. Our model performs better than related methods on the task of predicting traits responding to anticoagulants, in particular the blood clot formation trait PTT.

While predictive performance is similar across related methods, our method
shows important advantages in model flexibility. We demonstrate this through two case
studies: First, we study predictive trajectories and the inferred effects on blood pressure and heart rate of a treatment on one patient (Fig. \ref{fig:metoprolol_results}). The patient received 50
mg of metoprolol tartrate . We compare the predictive trajectory with
uncertainty from a related method (Mat32+LTI) (Fig. \ref{fig:metoprolol_results} a–b) with results from our method (Fig.
\ref{fig:metoprolol_results} c–d). Our method estimates an explicit treatment-induced latent force with strong
effects on both heart rate and systolic blood pressure (Fig. \ref{fig:metoprolol_results} e) that matches the time frame prescribed by clinical medicine. Indeed, the direction of estimated effects from the related method on SBP (Fig. \ref{fig:metoprolol_results} a and b) is the opposite of the clinical usage; this error may be due to delayed effect of the drug on SBP. While for this patient
the MAE of the related method is slightly lower than our method (4.68 vs. 5.37
in SBP; 7.90 vs. 8.09 in HR), results from our method are closer to clinical
ground truth with substantially lower uncertainty. Furthermore, with the GP model of latent force, our estimated effect is more robust than a standard latent force model when presented with additional uncertainty and noise in the data, such as delayed effects from the time of treatment administration, which is important in the clinical setting.

\begin{figure*}[t!]
    \centering
    \includegraphics[width=11cm]{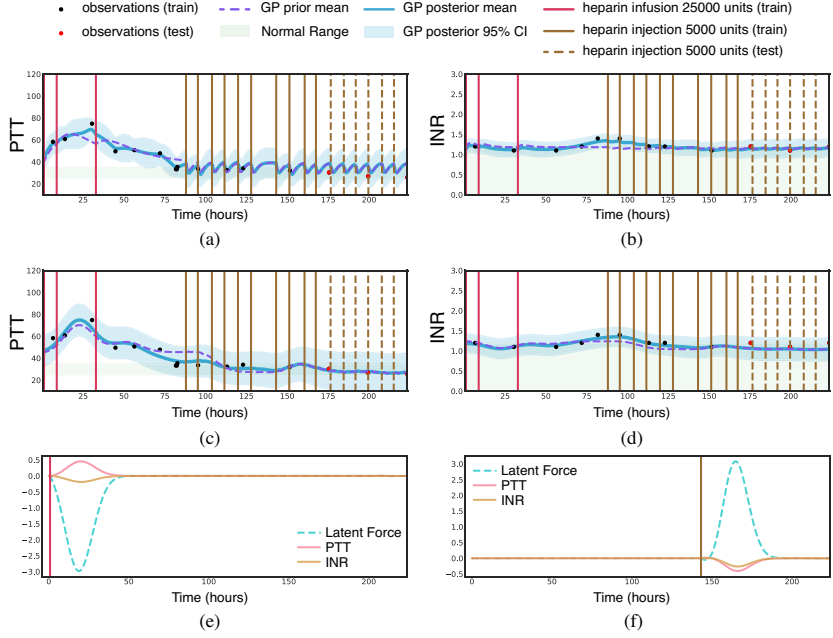}
    \caption{\textbf{Prediction of partial thromboplastin time (PTT) and international normalized ratio (INR) for one patient under heparin treatments.} (a) and (b): results from the related method \textsc{Mat32+Lti}. (c) and (d): results from the proposed method. (e) and (f): the estimated latent force for one heparin infusion (25000 units) and one heparin injection (5000 units), and their effects on PTT and INR; the recovered effects on PTT and INR are in their \emph{normalized} units.}
    \label{fig:heparin_results}
    \vspace{-2mm}
\end{figure*}

Next, we find similar robust and clinically interpretable behavior are achieved for a patient receiving multiple heparin infusions (25000 units) and heparin injections (5000 units; Fig. \ref{fig:heparin_results}). For both our method and the related method (\textsc{Mat32+Lti}), the longer-term effects on PTT is estimated (Fig. \ref{fig:heparin_results}, a and c). Both methods estimated negative effects for the heparin injections (Fig. \ref{fig:heparin_results} f), while in general heparin is assumed not to affect INR  \citep{katzung2015basic}.

\vspace{-1mm}
\section{Discussion}

We developed a framework using latent force models (LFMs) to capture treatment effects on patients' physiological state estimated using medical time series data. 
By modeling treatment effects as latent forces convolved with a multi-output GP, we create a flexible framework that bridges the gap between the smooth, stationary dynamics of patient state and a mechanistic understanding of the forces that impact clinical traits. A GP model of latent forces provides a flexible probabilistic framework with convenient inference properties; we enforce appropriate effect dynamics using a causal kernel. Two key contributions, the latent force model convolution and the associated causal kernel, lead to a computationally tractable solution with low variance and clinically-relevant interpretation of personalized treatment effects. Further improvements in speed may be found by adapting a recent kernel approximation \citep{guarnizo2018fast}.

There are several directions to improve our method for clinical treatment effect estimation. Our framework assumes that the effect of each treatment is independent of any other, and interactions between treatments are not modeled. These interactions could be modeled by modifying the kernel for the latent force component. In addition, the method assumes the decay parameters ($D_{i,j}$) of the treatment effect for a single patient are treatment-independent and constant throughout the patient's hospital stay.
As the decay parameters reflect the physiology of drug absorption, which may change as a function of patient state, we might model this parameter as a stochastic process itself.
For future research, the latent force model encourages an optimal control perspective: estimating treatment effect sizes of each patient for each clinical covariate, coupled with accurate patient state prediction, can provide a path toward treatment prioritization and decision-making in clinical interventions.

\bibliographystyle{apalike}
\bibliography{main}

\end{document}